\documentclass[10pt,twocolumn,letterpaper]{article}
\pdfoutput=1

\usepackage{cvpr}
\usepackage{times}
\usepackage{epsfig}
\usepackage{graphicx}
\usepackage{amsmath}
\usepackage{amssymb}
\usepackage{booktabs}
\usepackage{multirow,array}
\usepackage{color}
\usepackage{bm}
\usepackage{subcaption}
\usepackage{ifthen}
\usepackage[sort]{cite}       %
\usepackage[font=footnotesize,subrefformat=parens]{subcaption}
\usepackage{tabu}           %
\usepackage{xcolor}            %
\usepackage{graphicx}
\usepackage{tablefootnote}
\usepackage[pagebackref=true,breaklinks=true,letterpaper=true,colorlinks,bookmarks=false]{hyperref}

\definecolor{Plum}{RGB}{142, 69, 133}
\definecolor{Cyan}{RGB}{0, 255, 255} %
\definecolor{Red3}{HTML}{a40000}
\definecolor{Green3}{HTML}{4e9a06} %

\definecolor{tabutter}{rgb}{0.98824, 0.91373, 0.30980}      %
\definecolor{ta2butter}{rgb}{0.92941, 0.83137, 0}     %
\definecolor{ta3butter}{rgb}{0.76863, 0.62745, 0}     %

\definecolor{taorange}{rgb}{0.98824, 0.68627, 0.24314}      %
\definecolor{ta2orange}{rgb}{0.96078, 0.47451, 0}     %
\definecolor{ta3orange}{rgb}{0.80784, 0.36078, 0}     %

\definecolor{tachocolate}{rgb}{0.91373, 0.72549, 0.43137}   %
\definecolor{ta2chocolate}{rgb}{0.75686, 0.49020, 0.066667} %
\definecolor{ta3chocolate}{rgb}{0.56078, 0.34902, 0.0078431}   %

\definecolor{tachameleon}{rgb}{0.54118, 0.88627, 0.20392}   %
\definecolor{ta2chameleon}{rgb}{0.45098, 0.82353, 0.086275} %
\definecolor{ta3chameleon}{rgb}{0.30588, 0.60392, 0.023529} %

\definecolor{taskyblue}{rgb}{0.44706, 0.56078, 0.81176}     %
\definecolor{ta2skyblue}{rgb}{0.20392, 0.39608, 0.64314} %
\definecolor{ta3skyblue}{rgb}{0.12549, 0.29020, 0.52941} %

\definecolor{taplum}{rgb}{0.67843, 0.49804, 0.65882}     %
\definecolor{ta2plum}{rgb}{0.45882, 0.31373, 0.48235}    %
\definecolor{ta3plum}{rgb}{0.36078, 0.20784, 0.4}     %

\definecolor{tascarletred}{rgb}{0.93725, 0.16078, 0.16078}  %
\definecolor{ta2scarletred}{rgb}{0.8, 0, 0}        %
\definecolor{ta3scarletred}{rgb}{0.64314, 0, 0}       %

\definecolor{taaluminium}{rgb}{0.93333, 0.93333, 0.92549}   %
\definecolor{ta2aluminium}{rgb}{0.82745, 0.84314, 0.81176}  %
\definecolor{ta3aluminium}{rgb}{0.72941, 0.74118, 0.71373}  %

\definecolor{tagray}{rgb}{0.53333, 0.54118, 0.52157}     %
\definecolor{ta2gray}{rgb}{0.33333, 0.34118, 0.32549}    %
\definecolor{ta3gray}{rgb}{0.18039, 0.20392, 0.21176}    %

\newcommand{\reffig}[1]{Figure~\ref{fig:#1}}
\newcommand{\shortreffig}[1]{Fig.~\ref{fig:#1}}
\newcommand{\refsec}[1]{Section~\ref{sec:#1}}

\newcommand{\reftbl}[1]{Table~\ref{tbl:#1}}

\newcommand{\refeq}[1]{Equation~\eqref{eq:#1}}

\newcommand{\lblfig}[1]{\label{fig:#1}}
\newcommand{\lblsec}[1]{\label{sec:#1}}
\newcommand{\lbleq}[1]{\label{eq:#1}}
\newcommand{\lbltbl}[1]{\label{tbl:#1}}

\definecolor{lightpink}{rgb}{1.0, 0.71, 0.76}
\definecolor{wildwatermelon}{rgb}{0.99, 0.42, 0.52}

\newcommand\blfootnote[1]{%
\begingroup
\renewcommand\thefootnote{}\footnote{#1}%
\addtocounter{footnote}{-1}%
\endgroup
}

\cvprfinalcopy %

\def\cvprPaperID{1039} %

\ifcvprfinal\fi
\begin{document}

\title{Context Encoders: Feature Learning by Inpainting}

\author{Deepak Pathak
\and
Philipp Kr\"ahenb\"uhl
\and
Jeff Donahue
\and
Trevor Darrell
\and
Alexei A. Efros\\
\and
University of California, Berkeley\\
{\tt\small \{pathak,philkr,jdonahue,trevor,efros\}@cs.berkeley.edu}
}

\maketitle
\begin{abstract}
We present an unsupervised visual feature learning algorithm driven by context-based pixel prediction.
By analogy with auto-encoders, we propose Context Encoders -- a convolutional neural network trained to generate the contents of an arbitrary image region conditioned on its surroundings.
In order to succeed at this task, context encoders need to both understand the content of the entire image, as well as produce a plausible hypothesis for the missing part(s).
When training context encoders, we have experimented with
both a standard pixel-wise reconstruction loss, as well as a reconstruction plus an adversarial loss.  The latter produces much sharper results because it can better handle multiple modes in the output.
We found that a context encoder learns a representation that captures not just appearance but also the semantics of visual structures.
We quantitatively demonstrate the effectiveness of our learned features for CNN pre-training on classification, detection, and segmentation tasks.
Furthermore, context encoders can be used for semantic inpainting tasks, either stand-alone or as initialization for non-parametric methods.

\end{abstract}

\blfootnote{The code, trained models and more inpainting results are available at the author's project website.}

\section{Introduction}
Our visual world is very diverse, yet highly structured, and
humans have an uncanny ability to make sense of this structure.
In this work, we explore whether state-of-the-art computer vision algorithms can do the same.
Consider the image shown in~\reffig{teaser_input}.
Although the center part of the image is missing, most of us can easily imagine its content from the surrounding pixels, without having ever seen that exact scene. Some of us can even draw it, as shown on~\reffig{teaser_artist}.
This ability comes from the fact that natural images, despite their diversity, are highly structured (e.g. the regular pattern of windows on the facade).
We humans are able to understand this structure and make visual predictions even when seeing only parts of the scene.
In this paper, we show that it is possible to learn and predict this structure using convolutional neural networks (CNNs), a class of models that have recently shown success across a variety of image understanding tasks.
\begin{figure}
\vspace{-.8em}
\captionsetup[subfigure]{justification=centering}
  \begin{subfigure}[b]{.49\columnwidth}
    \centering
    \includegraphics[width=0.99\linewidth,height=0.99\linewidth]{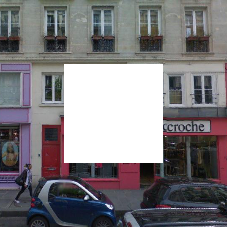}
    \caption{Input context}
    \lblfig{teaser_input}
  \end{subfigure}
  \begin{subfigure}[b]{.49\columnwidth}
    \centering
    \includegraphics[width=0.99\linewidth,height=0.99\linewidth]{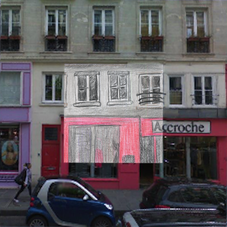}
    \caption{Human artist}
    \lblfig{teaser_artist}
  \end{subfigure}
  \vspace{0.8mm}

  \begin{subfigure}[b]{.49\columnwidth}
    \centering
    \includegraphics[width=0.99\linewidth,height=0.99\linewidth]{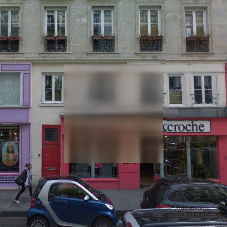}
    \caption{Context Encoder\\($L2$ loss)}
    \lblfig{teaser_rec}
  \end{subfigure}
  \begin{subfigure}[b]{.49\columnwidth}
    \centering
    \includegraphics[width=0.99\linewidth,height=0.99\linewidth]{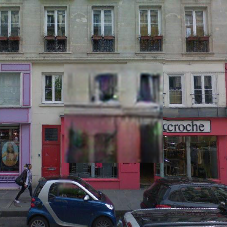}
    \caption{Context Encoder\\($L2$ + Adversarial loss)}
    \lblfig{teaser_adv}
  \end{subfigure}
\vspace{-0.5em}
  \caption{Qualitative illustration of the task. Given an image with a missing region (a), a human artist has no trouble inpainting it (b).  Automatic inpainting using our \textit{context encoder} trained with $L2$ reconstruction loss is shown in (c), and using both $L2$ and adversarial losses in (d).}
  \lblfig{teaser}
\vspace{-.6em}
\end{figure}

Given an image with a missing region (e.g., \shortreffig{teaser_input}), we train a convolutional neural network to regress to the missing pixel values (\shortreffig{teaser_adv}).
We call our model \textit{context encoder}, as it consists of an encoder capturing the context of an image into a compact latent feature representation and a decoder which uses that representation to produce the missing image content.
The context encoder is closely related to autoencoders~\cite{hintonautoenc,bengio2009learning}, as it shares a similar encoder-decoder architecture.
Autoencoders take an input image and try to reconstruct it after it passes through a low-dimensional ``bottleneck'' layer, with the aim of obtaining a compact feature representation of the scene.
Unfortunately, this feature representation is likely to just compresses the image content without learning a semantically meaningful representation.
Denoising autoencoders~\cite{denoising} address this issue by corrupting the input image and requiring the network to undo the damage.
However, this corruption process is typically very localized and low-level, and does not require much semantic information to undo.
In contrast, our context encoder needs to solve a much harder task: to fill in large missing areas of the image, where it can't get ``hints'' from nearby pixels.
This requires a much deeper semantic understanding of the scene, and the ability to synthesize high-level features over large spatial extents.
For example, in~\reffig{teaser_input}, an entire window needs to be conjured up ``out of thin air.''  This is similar in spirit to word2vec~\cite{mikolov2013distributed} which learns word representation from natural language sentences by predicting a word given its context.

Like autoencoders, context encoders are trained in a completely unsupervised manner.
Our results demonstrate that in order to succeed at this task, a model needs to both understand the content of an image, as well as produce a plausible hypothesis for the missing parts.
This task, however, is inherently multi-modal as there are multiple ways to fill the missing region while also maintaining coherence with the given context.
We decouple this burden in our loss function by jointly training our context encoders to minimize both a reconstruction loss and an adversarial loss.
The reconstruction (L2) loss captures the overall structure of the missing region in relation to the context, while the the adversarial loss~\cite{goodfellow2014generative}
has the effect of picking a particular mode from the distribution.  \reffig{teaser} shows that using only the reconstruction loss produces blurry results, whereas adding the adversarial loss results in much sharper predictions.

We evaluate the encoder and the decoder independently.
On the encoder side, we show that encoding just the context of an image patch and using the resulting feature to retrieve nearest neighbor contexts from a dataset produces patches which are semantically similar to the original (unseen) patch.
We further validate the quality of the learned feature representation by fine-tuning the encoder for a variety of image understanding tasks, including classification, object detection, and semantic segmentation. We are competitive
with the state-of-the-art unsupervised/self-supervised methods on those tasks.
On the decoder side, we show that our method is often able to fill in realistic image content.
Indeed, to the best of our knowledge, ours is the first parametric inpainting algorithm that is able to give reasonable results for semantic hole-filling (i.e. large missing regions).  The context encoder can also be useful as a better visual feature for computing nearest neighbors in non-parametric inpainting methods.

\section{Related work}
Computer vision has made tremendous progress on semantic image understanding tasks such as classification, object detection, and segmentation in the past decade.
Recently, Convolutional Neural Networks (CNNs)~\cite{fukushima1980neocognitron,lecun1989backpropagation} have greatly advanced the performance in these tasks~\cite{krizhevsky2012imagenet,rcnn,long2014fully}.
The success of such models on image classification paved the way to tackle harder problems, including unsupervised understanding and generation of natural images.
We briefly review the related work in each of the sub-fields pertaining to this paper.

\paragraph{Unsupervised learning}
CNNs trained for ImageNet~\cite{imagenet} classification with over a million labeled examples learn features which generalize very well across tasks~\cite{donahue2013decaf}.
However, whether such semantically informative and generalizable features can be learned from raw images alone, without any labels, remains an open question.
Some of the earliest work in deep unsupervised learning are autoencoders~\cite{hintonautoenc,bengio2009learning}.
Along similar lines, denoising autoencoders~\cite{denoising} reconstruct the image from local corruptions, to make encoding robust to such corruptions.
While context encoders could be thought of as a variant of denoising autoencoders, the corruption applied to the model's input is spatially much larger, requiring more semantic information to undo.

\paragraph{Weakly-supervised and self-supervised learning}
Very recently, there has been significant interest in learning meaningful representations using weakly-supervised and self-supervised learning.
One useful source of supervision is to use the temporal information contained in videos.
Consistency across temporal frames has been used as supervision to learn embeddings which perform well on a number of tasks~\cite{goroshin2015unsupervised,ramanathan2015learning}.
Another way to use consistency is to track patches in frames of video containing task-relevant attributes and use the coherence of tracked patches to guide the training~\cite{wang2015unsupervised}.
Ego-motion read off from non-vision sensors has been used as supervisory signal to train visual features~\etal~\cite{agrawal2015learning,jayaraman2015learning}.

Most closely related to the present paper are efforts at exploiting spatial context as a source of free and plentiful supervisory signal.
Visual Memex~\cite{malisiewicz2009beyond} used context to non-parametrically model object relations and to predict masked objects in scenes, while~\cite{doersch2014context} used context to establish correspondences for unsupervised object discovery.
However, both approaches relied on hand-designed features and did not perform any representation learning.
Recently, Doersch~\etal~\cite{doersch2015unsupervised} used the task of predicting the relative positions of neighboring patches within an image as a way to train an unsupervised deep feature representations.  We share the same high-level goals with Doersch~\etal but fundamentally differ in the approach: whereas \cite{doersch2015unsupervised} are solving a \textit{discriminative} task (is patch A above patch B or below?), our context encoder solves a pure \textit{prediction} problem (what pixel intensities should go in the hole?).  Interestingly, similar distinction exist in using language context to learn word embeddings: Collobert and Weston~\cite{collobert2008unified} advocate a discriminative approach, whereas word2vec~\cite{mikolov2013distributed} formulate it as word prediction.
One important benefit of our approach is that our supervisory signal is much richer: a context encoder needs to predict roughly 15,000 real values per training example, compared to just 1 option among 8 choices in \cite{doersch2015unsupervised}.
Likely due in part to this difference, our context encoders take far less time to train than~\cite{doersch2015unsupervised}.
Moreover, context based prediction is also harder to ``cheat'' since low-level image features, such as chromatic aberration, do not provide any meaningful information, in contrast to~\cite{doersch2015unsupervised} where chromatic aberration partially solves the task.  On the other hand, it is not yet clear if requiring faithful pixel generation is necessary for learning good visual features.

\paragraph{Image generation}
Generative models of natural images have enjoyed significant research interest~\cite{ranzato2013modeling,kingma2013auto,goodfellow2014generative}.
Recently, Radford~\etal~\cite{dcgan} proposed new convolutional architectures and optimization hyperparameters for Generative Adversarial Networks (GAN)~\cite{goodfellow2014generative} producing encouraging results.
We train our context encoders using an adversary jointly with reconstruction loss for generating inpainting results.
We discuss this in detail in Section~\ref{sec:loss}.

Dosovitskiy~\etal~\cite{chairs} and Rifai~\etal~\cite{torontofaces} demonstrate that CNNs can learn to generate novel images of particular object categories (chairs and faces, respectively), but rely on large labeled datasets with examples of these categories.
In contrast, context encoders can be applied to any unlabeled image database and learn to generate images based on the surrounding context.

\paragraph{Inpainting and hole-filling}
It is important to point out that our hole-filling task cannot be handled by classical inpainting~\cite{bertalmio2000image,osher2005iterative} or texture synthesis ~\cite{efros1999texture,barnes2009patchmatch} approaches, since the missing region is too large for local non-semantic methods to work well.
In computer graphics, filling in large holes is typically done via scene completion~\cite{hays2007scene}, involving a cut-paste formulation using nearest neighbors from a dataset of millions of images. However, scene completion is meant for filling in holes left by removing whole objects, and it struggles to fill arbitrary holes, e.g. amodal completion of partially occluded objects.
Furthermore, previous completion relies on a hand-crafted distance metric, such as Gist~\cite{gist} for nearest-neighbor computation which is inferior to a learned distance metric.
We show that our method is often able to inpaint semantically meaningful content in a parametric fashion, as well as provide a better feature for nearest neighbor-based inpainting methods.

\section{Context encoders for image generation}
We now introduce context encoders: CNNs that predict missing parts of a scene from their surroundings.
We first give an overview of the general architecture, then provide details on the learning procedure and finally present various strategies for image region removal.

\subsection{Encoder-decoder pipeline}
\begin{figure}[t]
\includegraphics[width=\linewidth]{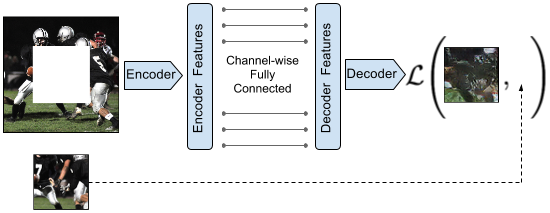}
\vspace{-0.5em}
\caption{Context Encoder. The context image is passed through the encoder to obtain features which are connected to the decoder using channel-wise fully-connected layer as described in~\refsec{method}. The decoder then produces the missing regions in the image.}
\lblfig{concept}
\vspace{-0.5em}
\end{figure}
\lblsec{method}
The overall architecture is a simple encoder-decoder pipeline.
The encoder takes an input image with missing regions and produces a latent feature representation of that image.
The decoder takes this feature representation and produces the missing image content.
We found it important to connect the encoder and the decoder through a channel-wise fully-connected layer, which allows each unit in the decoder to reason about the entire image content.
\reffig{concept} shows an overview of our architecture.

\paragraph{Encoder}
Our encoder is derived from the \textit{AlexNet} architecture~\cite{krizhevsky2012imagenet}.
Given an input image of size $227\times 227$, we use the first five convolutional layers and the following pooling layer (called \textit{pool5}) to compute an abstract $6 \times 6 \times 256$ dimensional feature representation.
In contrast to \textit{AlexNet}, our model is not trained for ImageNet classification; rather, the network is trained for context prediction ``from scratch'' with randomly initialized weights.

However, if the encoder architecture is limited only to convolutional layers, there is no way for information to directly propagate from one corner of the feature map to another.
This is so because convolutional layers connect all the feature maps together, but never directly connect all locations within a specific feature map.
In the present architectures, this information propagation is handled by \textit{fully-connected} or \textit{inner product} layers, where all the activations are directly connected to each other.
In our architecture, the latent feature dimension is $6\times 6 \times 256 = 9216$ for both encoder and decoder.
This is so because, unlike autoencoders, we do not reconstruct the original input and hence need not have a smaller \textit{bottleneck}.
However, fully connecting the encoder and decoder would result in an explosion in the number of parameters (over 100M!), to the extent that efficient training on current GPUs would be difficult.
To alleviate this issue, we use a channel-wise fully-connected layer to connect the encoder features to the decoder, described in detail below.

\paragraph{Channel-wise fully-connected layer}
This layer is essentially a fully-connected layer with groups, intended to propagate information within activations of each feature map.
If the input layer has $m$ feature maps of size $n\times n$, this layer will output $m$ feature maps of dimension $n\times n$.
However, unlike a fully-connected layer, it has no parameters connecting different feature maps and only propagates information within feature maps.
Thus, the number of parameters in this channel-wise fully-connected layer is $mn^4$, compared to $m^2n^4$ parameters in a fully-connected layer (ignoring the bias term).
This is followed by a stride 1 convolution to propagate information across channels.

\paragraph{Decoder}
We now discuss the second half of our pipeline, the decoder, which generates pixels of the image using the encoder features.
The ``encoder features'' are connected to the ``decoder features'' using a channel-wise fully-connected layer.

The channel-wise fully-connected layer is followed by a series of five \textit{up-convolutional} layers~\cite{ZeilerF14,chairs,long2014fully} with learned filters, each with a rectified linear unit (ReLU) activation function.
A up-convolutional is simply a convolution that results in a higher resolution image.
It can be understood as upsampling followed by convolution (as described in~\cite{chairs}), or convolution with fractional stride (as described in~\cite{long2014fully}).
The intuition behind this is straightforward -- the series of up-convolutions and non-linearities comprises a non-linear weighted upsampling of the feature produced by the encoder until we roughly reach the original target size.

\subsection{Loss function}
\label{sec:loss}
We train our context encoders by regressing to the ground truth content of the missing (dropped out) region.
However, there are often multiple equally plausible ways to fill a missing image region which are consistent with the context.
We model this behavior by having a decoupled joint loss function to handle both continuity within the context and multiple modes in the output.
The reconstruction (L2) loss is responsible for capturing the overall structure of the missing region and coherence with regards to its context, but tends to average together the multiple modes in predictions.
The adversarial loss~\cite{goodfellow2014generative}, on the other hand,  tries to make prediction look real, and has the effect of picking a particular mode from the distribution.
For each ground truth image $x$, our context encoder $F$ produces an output $F(x)$.
Let $\hat{M}$ be a binary mask corresponding to the dropped image region with a value of 1 wherever a pixel was dropped and 0 for input pixels.
During training, those masks are automatically generated for each image and training iterations, as described in \refsec{regdrop}.
We now describe different components of our loss function.

\paragraph{Reconstruction Loss}
We use a normalized masked $L2$ distance as our reconstruction loss function, $\mathcal{L}_{rec}$,
\begin{align}
\mathcal{L}_{rec}(x) = \|\hat{M}\odot(x-F((1-\hat{M})\odot x))\|_2^2 \lbleq{loss_rec},
\end{align}
where $\odot$ is the element-wise product operation.
We experimented with both L1 and L2 losses and found no significant difference between them.
While this simple loss encourages the decoder to produce a rough outline of the predicted object, it often fails to capture any high frequency detail (see~\shortreffig{teaser_rec}).
This stems from the fact that the L2 (or L1) loss often prefer a blurry  solution, over highly accurate textures.
We believe this happens because it is much ``safer'' for the L2 loss to predict the mean of the distribution, because this minimizes the mean pixel-wise error, but results in a blurry averaged image.
We alleviated this problem by adding an adversarial loss.

\begin{figure}[t]
\vspace{-0.5em}
\centering
\begin{subfigure}[b]{0.32\linewidth}
\centering
\includegraphics[width=0.7\linewidth]{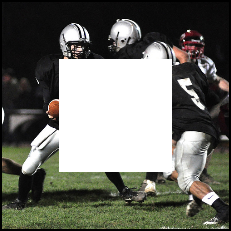}
\includegraphics[width=0.7\linewidth]{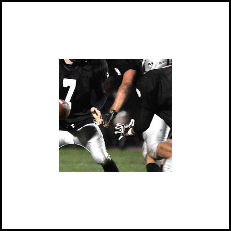}
\caption{Central region}
\lblfig{drop_random}
\end{subfigure}
\begin{subfigure}[b]{0.32\linewidth}
\centering
\includegraphics[width=0.7\linewidth]{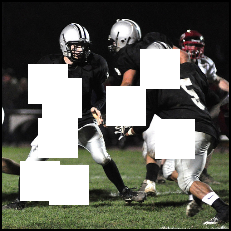}
\includegraphics[width=0.7\linewidth]{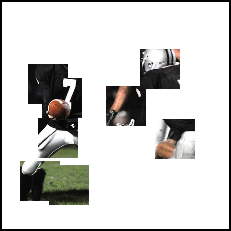}
\caption{Random block}
\lblfig{drop_block}
\end{subfigure}
\begin{subfigure}[b]{0.32\linewidth}
\centering
\includegraphics[width=0.7\linewidth]{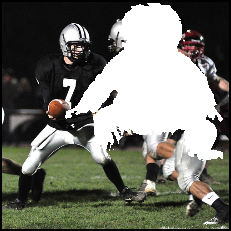}
\includegraphics[width=0.7\linewidth]{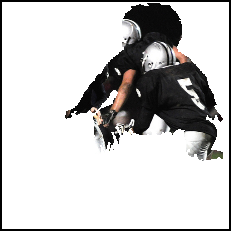}
\caption{Random region}
\lblfig{drop_region}
\end{subfigure}
\vspace{-0.5em}
\caption{An example of image $x$ with our different region masks $\hat M$ applied, as described in Section~\ref{sec:regdrop}.}
\label{fig:regdrop}
\vspace{-0.5em}
\end{figure}

\begin{figure*}[t]
\vspace{-0.5em}
\centering
\begin{tabular}{c@{\hskip 1pt}c@{\hskip 4pt}c@{\hskip 1pt}c@{\hskip 4pt}c@{\hskip 1pt}c@{\hskip 4pt}c@{\hskip 1pt}c}
\includegraphics[width=0.12\linewidth]{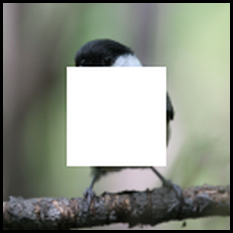} &
\includegraphics[width=0.12\linewidth]{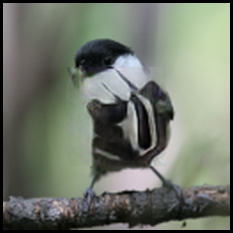} &
\includegraphics[width=0.12\linewidth]{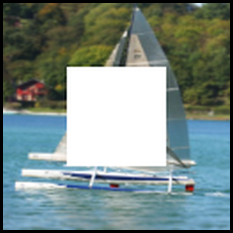} &
\includegraphics[width=0.12\linewidth]{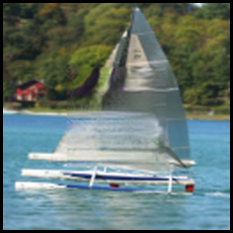} &
\includegraphics[width=0.12\linewidth]{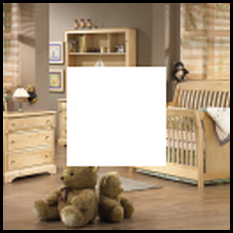} &
\includegraphics[width=0.12\linewidth]{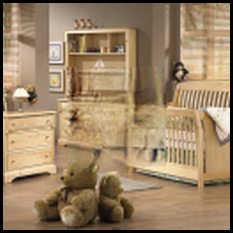} &
\includegraphics[width=0.12\linewidth]{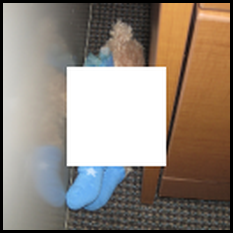} &
\includegraphics[width=0.12\linewidth]{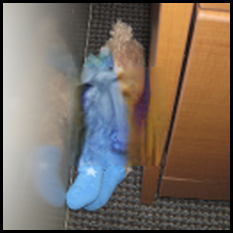} \\
\includegraphics[width=0.12\linewidth]{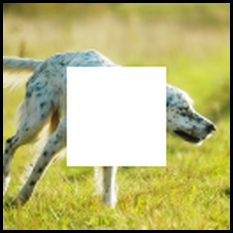} &
\includegraphics[width=0.12\linewidth]{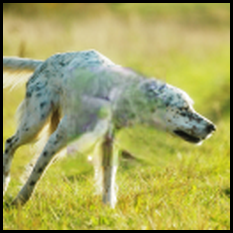} &
\includegraphics[width=0.12\linewidth]{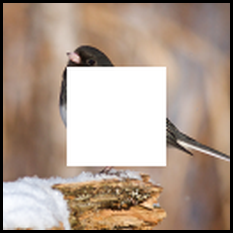} &
\includegraphics[width=0.12\linewidth]{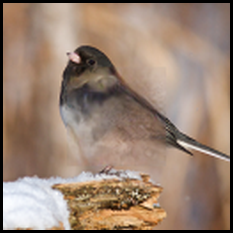} &
\includegraphics[width=0.12\linewidth]{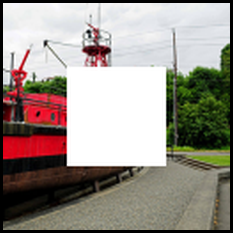} &
\includegraphics[width=0.12\linewidth]{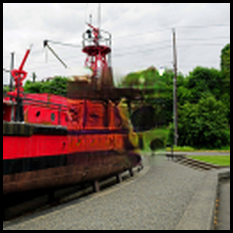} &
\includegraphics[width=0.12\linewidth]{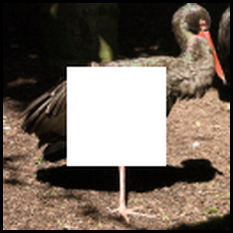} &
\includegraphics[width=0.12\linewidth]{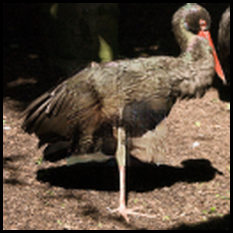} \\
\includegraphics[width=0.12\linewidth]{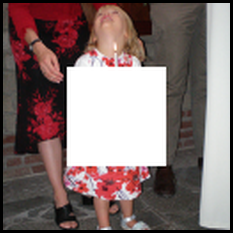} &
\includegraphics[width=0.12\linewidth]{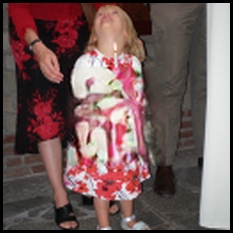} &
\includegraphics[width=0.12\linewidth]{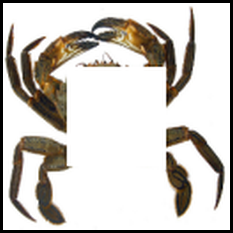} &
\includegraphics[width=0.12\linewidth]{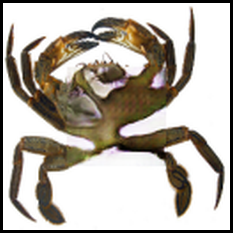} &
\includegraphics[width=0.12\linewidth]{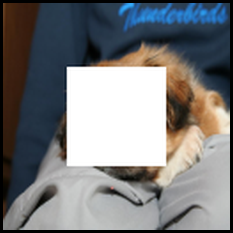} &
\includegraphics[width=0.12\linewidth]{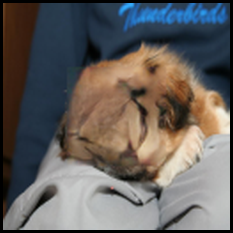} &
\includegraphics[width=0.12\linewidth]{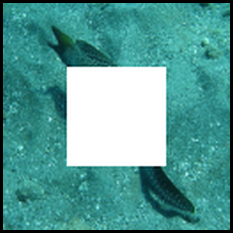} &
\includegraphics[width=0.12\linewidth]{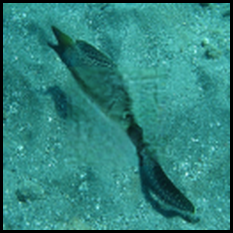} \\
\includegraphics[width=0.12\linewidth]{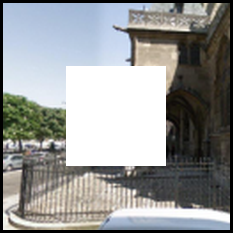} &
\includegraphics[width=0.12\linewidth]{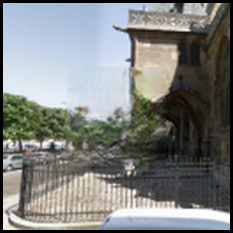} &
\includegraphics[width=0.12\linewidth]{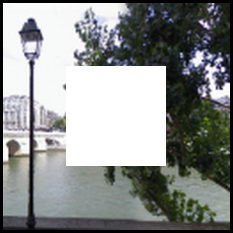} &
\includegraphics[width=0.12\linewidth]{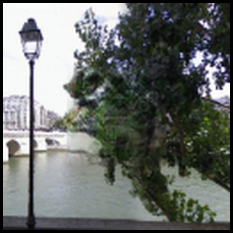} &
\includegraphics[width=0.12\linewidth]{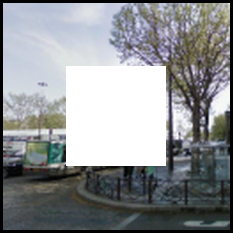} &
\includegraphics[width=0.12\linewidth]{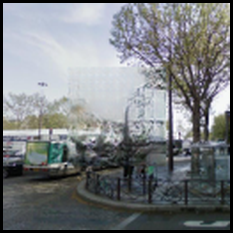} &
\includegraphics[width=0.12\linewidth]{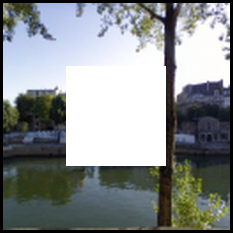} &
\includegraphics[width=0.12\linewidth]{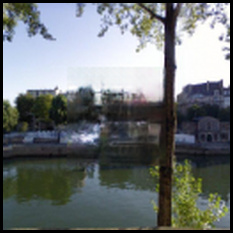} \\
\includegraphics[width=0.12\linewidth]{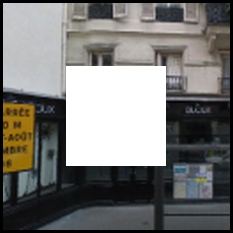} &
\includegraphics[width=0.12\linewidth]{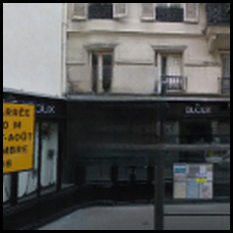} &
\includegraphics[width=0.12\linewidth]{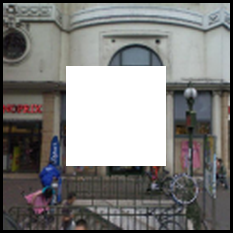} &
\includegraphics[width=0.12\linewidth]{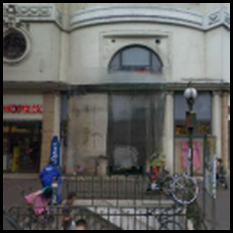} &
\includegraphics[width=0.12\linewidth]{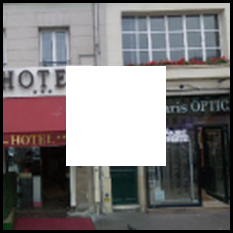} &
\includegraphics[width=0.12\linewidth]{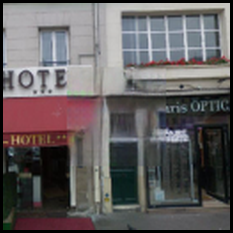} &
\includegraphics[width=0.12\linewidth]{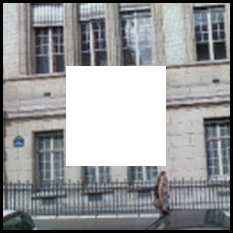} &
\includegraphics[width=0.12\linewidth]{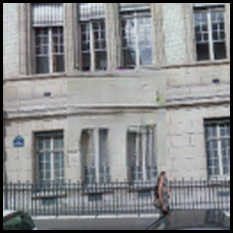}
\end{tabular}
\vspace{-0.5em}
\caption{Semantic Inpainting results on \textit{held-out} images for context encoder trained using reconstruction and adversarial loss. First three rows are examples from ImageNet, and bottom two rows are from Paris StreetView Dataset. See more results on author's project website.}
\lblfig{good_results}
\vspace{-0.5em}
\end{figure*}

\paragraph{Adversarial Loss}
Our adversarial loss is based on Generative Adversarial Networks (GAN)~\cite{goodfellow2014generative}.
To learn a generative model $G$ of a data distribution, GAN proposes to jointly learn an adversarial discriminative model $D$ to provide loss gradients to the generative model.
$G$ and $D$ are parametric functions (e.g., deep networks) where $G:\mathcal{Z}\rightarrow\mathcal{X}$ maps samples from noise distribution $\mathcal{Z}$ to data distribution $\mathcal{X}$.
The learning procedure is a two-player game where an adversarial discriminator $D$ takes in both the prediction of $G$ and ground truth samples, and tries to distinguish them, while $G$ tries to confuse $D$ by producing samples that appear as ``real'' as possible.
The objective for discriminator is logistic likelihood indicating whether the input is real sample or predicted one:
\begin{align}
\min_{G} \max_{D} & & & \mathbb{E}_{x\in \mathcal{X}}[\log (D(x))] + \mathbb{E}_{z\in \mathcal{Z}}[\log (1-D(G(z)))] \nonumber &
\end{align}

This method has recently shown encouraging results in generative modeling of images~\cite{dcgan}.
We thus adapt this framework for context prediction by modeling generator by context encoder; i.e., $G \triangleq F$.
To customize GANs for this task, one could condition on the given context information; i.e., the mask $\hat{M} \odot x$.
However, conditional GANs don't train easily for context prediction task as the adversarial discriminator $D$ easily exploits the perceptual discontinuity in generated regions and the original context to easily classify predicted versus real samples.
We thus use an alternate formulation, by conditioning only the generator (not the discriminator) on context.
We also found results improved when the generator was not conditioned on a noise vector.
Hence the adversarial loss for context encoders, $\mathcal{L}_{adv}$, is
\begin{align}
\mathcal{L}_{adv} = & \max_{D} & & \mathbb{E}_{x\in \mathcal{X}}\lbrack\log (D(x)) & \nonumber\\
& & & + \log (1-D(F((1-\hat{M})\odot x)))\rbrack, & \lbleq{loss_adv}
\end{align}
where, in practice, both $F$ and $D$ are optimized jointly using alternating SGD.
Note that this objective encourages the entire output of the context encoder to look realistic, not just the missing regions as in \refeq{loss_rec}.

\paragraph{Joint Loss}
We define the overall loss function as
\begin{align}
\mathcal{L} = \lambda_{rec}\mathcal{L}_{rec} + \lambda_{adv}\mathcal{L}_{adv} \lbleq{loss}.
\end{align}
Currently, we use adversarial loss only for inpainting experiments as AlexNet~\cite{krizhevsky2012imagenet} architecture training diverged with joint adversarial loss.
Details follow in Sections~\ref{sec:inpaint},~\ref{sec:feature}.

\subsection{Region masks}
\label{sec:regdrop}
The input to a context encoder is an image with one or more of its regions ``dropped out''; i.e., set to zero, assuming zero-centered inputs.
The removed regions could be of any shape, we present three different strategies here:

{\bf Central region} The simplest such shape is the central square patch in the image, as shown in \reffig{drop_random}.
While this works quite well for inpainting, the network learns low level image features that latch onto the boundary of the central mask.
Those low level image features tend not to generalize well to images without masks, hence the features learned are not very general.

{\bf Random block} To prevent the network from latching on the the constant boundary of the masked region, we randomize the masking process.
Instead of choosing a single large mask at a fixed location, we remove a number of smaller possibly overlapping masks, covering up to $\frac{1}{4}$ of the image.
An example of this is shown in \reffig{drop_block}.
However, the random block masking still has sharp boundaries convolutional features could latch onto.

{\bf Random region} To completely remove those boundaries, we experimented with removing arbitrary shapes from images, obtained from random masks in the PASCAL VOC 2012 dataset~\cite{everingham2014pascal}.
We deform those shapes and paste in arbitrary places in the other images (not from PASCAL), again covering up to $\frac{1}{4}$ of the image.
Note that we completely randomize the region masking process, and do not expect or want any correlation between the source segmentation mask and the image.
We merely use those regions to prevent the network from learning low-level features corresponding to the removed mask.
See example in \reffig{drop_region}.

In practice, we found region and random block masks produce a similarly general feature, while significantly outperforming the central region features.
We use the random region dropout for all our feature based experiments.

\begin{figure}[t]
\vspace{-0.5em}
\centering
\begin{tabular}{c@{\hskip 4pt}c@{\hskip 4pt}c}
\includegraphics[width=0.28\linewidth]{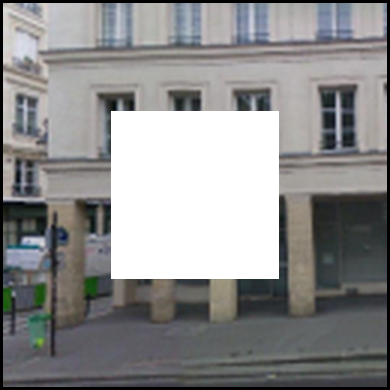} &
\includegraphics[width=0.28\linewidth]{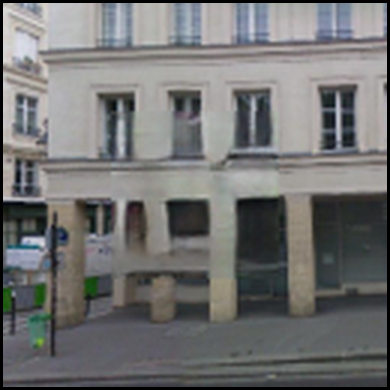} &
\includegraphics[width=0.28\linewidth]{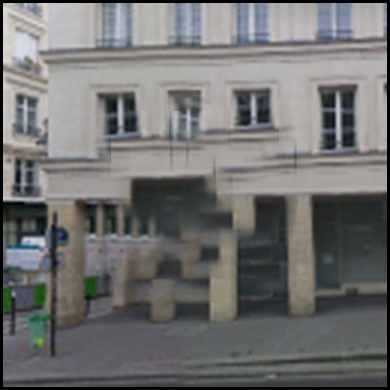} \\
\includegraphics[width=0.28\linewidth]{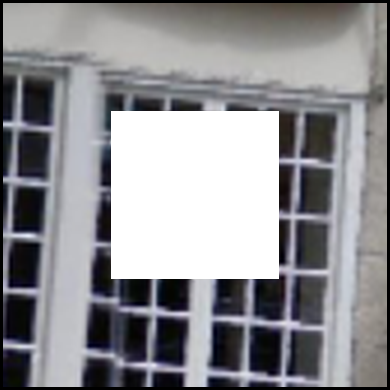} &
\includegraphics[width=0.28\linewidth]{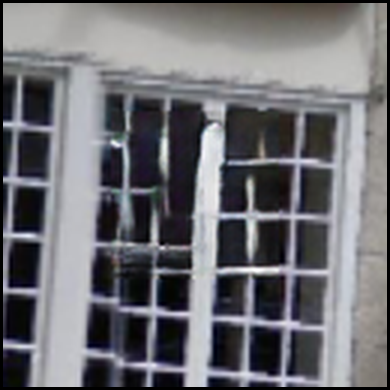} &
\includegraphics[width=0.28\linewidth]{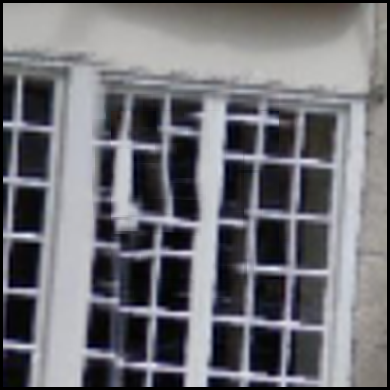} \\
\small Input Context & \small  Context Encoder & \small  Content-Aware Fill
\end{tabular}
\vspace{-0.5em}
\caption{Comparison with Content-Aware Fill (Photoshop feature based on~\cite{barnes2009patchmatch}) on \textit{held-out} images. Our method works better in semantic cases (top row) and works slightly worse in textured settings (bottom row).}
\lblfig{bad_results}
\vspace{-0.5em}
\end{figure}

\begin{figure*}[t]
\vspace{-0.5em}
\centering
\begin{tabular}{c@{\hskip 7pt}c@{\hskip 1pt}c@{\hskip 1pt}c@{\hskip 7pt}c@{\hskip 1pt}c@{\hskip 7pt}c@{\hskip 1pt}c}
\small Image & \small Ours($L2$) & \small Ours(Adv) & \small Ours(L2+Adv) & \multicolumn{2}{c}{\small NN-Inpainting w/ our features} & \multicolumn{2}{c}{\small NN-Inpainting w/ HOG}\\
\includegraphics[width=0.115\linewidth]{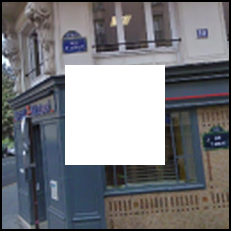} &
\includegraphics[width=0.115\linewidth]{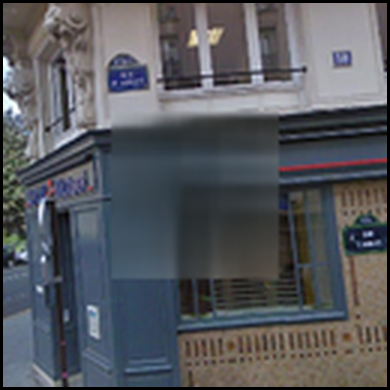} &
\includegraphics[width=0.115\linewidth]{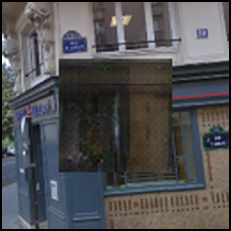} &
\includegraphics[width=0.115\linewidth]{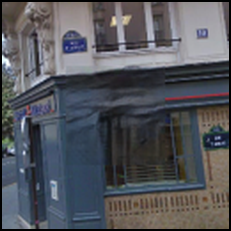} &
\includegraphics[width=0.115\linewidth]{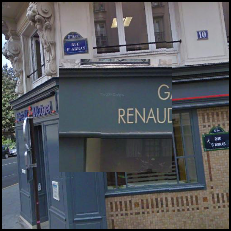} &
\includegraphics[width=0.115\linewidth]{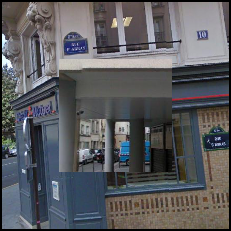} &
\includegraphics[width=0.115\linewidth]{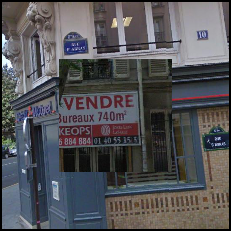} &
\includegraphics[width=0.115\linewidth]{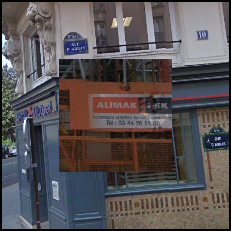} \\
\includegraphics[width=0.115\linewidth]{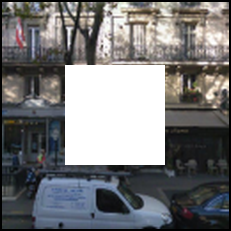} &
\includegraphics[width=0.115\linewidth]{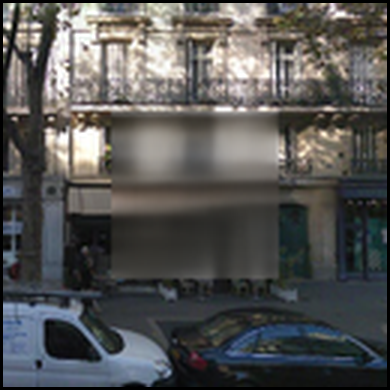} &
\includegraphics[width=0.115\linewidth]{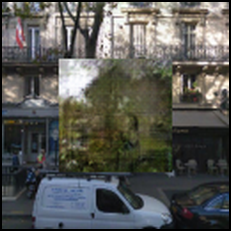} &
\includegraphics[width=0.115\linewidth]{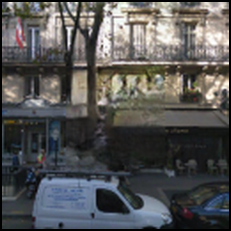} &
\includegraphics[width=0.115\linewidth]{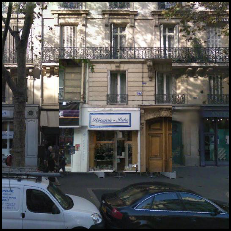} &
\includegraphics[width=0.115\linewidth]{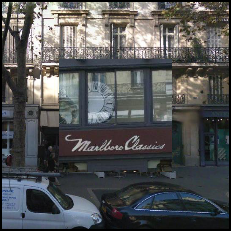} &
\includegraphics[width=0.115\linewidth]{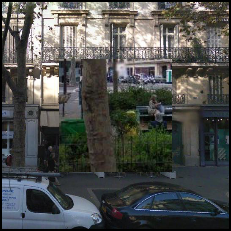} &
\includegraphics[width=0.115\linewidth]{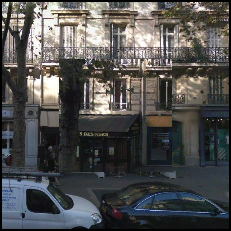} \\
\includegraphics[width=0.115\linewidth]{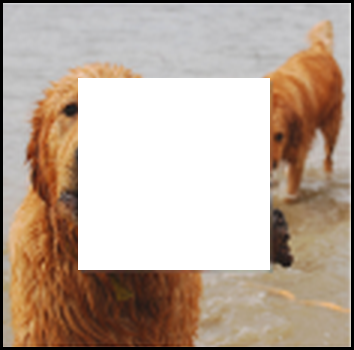} &
\includegraphics[width=0.115\linewidth]{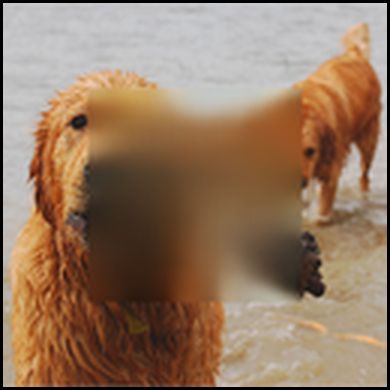} &
\includegraphics[width=0.115\linewidth]{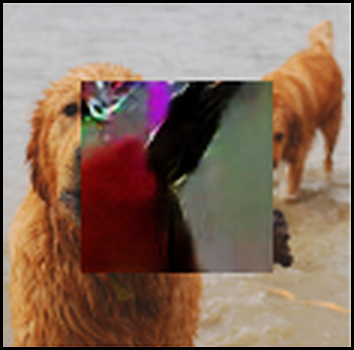} &
\includegraphics[width=0.115\linewidth]{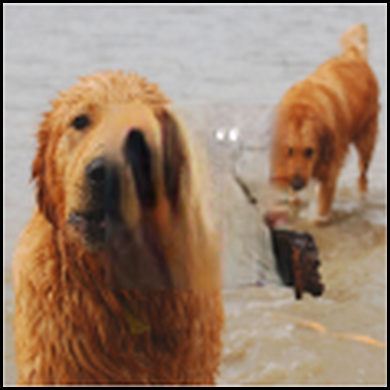} &
\includegraphics[width=0.115\linewidth]{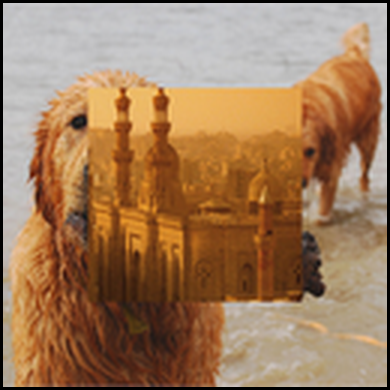} &
\includegraphics[width=0.115\linewidth]{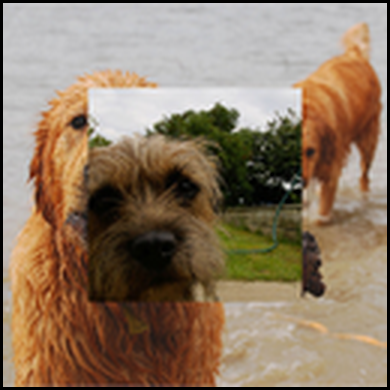} &
\includegraphics[width=0.115\linewidth]{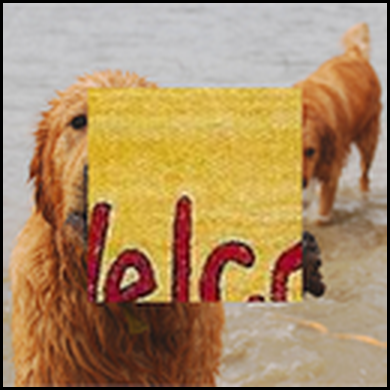} &
\includegraphics[width=0.115\linewidth]{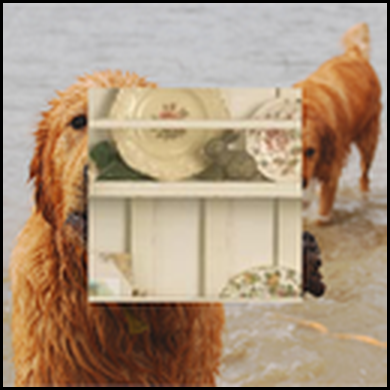}
\end{tabular}
\vspace{-0.5em}
\caption{Semantic Inpainting using different methods on \textit{held-out} images.
Context Encoder with just L2 are well aligned, but not sharp. Using adversarial loss, results are sharp but not coherent. Joint loss alleviate the weaknesses of each of them.
The last two columns are the results if we plug-in the best nearest neighbor (NN) patch in the masked region.
}
\lblfig{inpaint}
\vspace{-0.5em}
\end{figure*}

\section{Implementation details}
The pipeline was implemented in \textit{Caffe}~\cite{caffe} and Torch.
We used the recently proposed stochastic gradient descent solver, ADAM~\cite{kingma2014adam} for optimization.
The missing region in the masked input image is filled with constant mean value.
Hyper-parameter details are discussed in Sections~\ref{sec:inpaint},~\ref{sec:feature}.

{\bf Pool-free encoders}
We experimented with replacing all pooling layers with convolutions of the same kernel size and stride.
The overall stride of the network remains the same, but it results in finer inpainting.
Intuitively, there is no reason to use pooling for reconstruction based networks.
In classification, pooling provides spatial invariance, which may be detrimental for reconstruction-based training.
To be consistent with prior work, we still use the original AlexNet architecture (with pooling) for all feature learning results.

\section{Evaluation}
We now evaluate the encoder features for their semantic quality and transferability to other image understanding tasks.
We experiment with images from two datasets: Paris StreetView~\cite{doersch2012makes} and ImageNet~\cite{imagenet} without using any of the accompanying labels.
In Section~\ref{sec:inpaint}, we present visualizations demonstrating the ability of the context encoder to fill in semantic details of images with missing regions.
In Section~\ref{sec:feature}, we demonstrate the transferability of our learned features to other tasks, using context encoders as a pre-training step for image classification, object detection, and semantic segmentation.
We compare our results on these tasks with those of other unsupervised or self-supervised methods, demonstrating that our approach outperforms previous methods.

\begin{table}[t]
\centering
\resizebox{\linewidth}{!}{%
\begin{tabular}{lccc}
\toprule
\small Method &\small Mean L1 Loss &\small Mean L2 Loss &\small PSNR (higher better)\\
\midrule
\small NN-inpainting (HOG features) & 19.92\% & 6.92\% & 12.79 dB \\
\midrule
\small NN-inpainting (our features) & 15.10\% & 4.30\% & 14.70 dB \\
\small Our Reconstruction (joint) & \textbf{09.37\%} & \textbf{1.96\%} & \textbf{18.58 dB}\\
\bottomrule
\end{tabular}}
\vspace{-0.5em}
\caption{\small Semantic Inpainting accuracy for Paris StreetView dataset on \textit{held-out} images. NN inpainting is basis for~\cite{hays2007scene}.}
\lbltbl{inpaint_table}
\vspace{-0.5em}
\end{table}

\subsection{Semantic Inpainting}
\lblsec{inpaint}
We train context encoders with the joint loss function defined in Equation~\eqref{eq:loss} for the task of inpainting the missing region.
The encoder and discriminator architecture is similar to that of discriminator in~\cite{dcgan}, and decoder is similar to generator in~\cite{dcgan}.
However, the bottleneck is of $4000$ units (in contrast to $100$ in~\cite{dcgan}); see supplementary material.
We used the default solver hyper-parameters suggested in~\cite{dcgan}.
We use $\lambda_{rec}=0.999$ and $\lambda_{adv}=0.001$.
However, a few things were crucial for training the model. We did not condition the adversarial loss (see~\refsec{loss}) nor did we add noise to the encoder. We use a higher learning rate for context encoder (10 times) to that of adversarial discriminator.
To further emphasize the consistency of prediction with the context, we predict a slightly larger patch that overlaps with the context (by 7px). During training, we use higher weight ($10\times$) for the reconstruction loss in this overlapping region.

The qualitative results are shown in~\reffig{good_results}.
Our model performs generally well in inpainting semantic regions of an image.
However, if a region can be filled with low-level textures, texture synthesis methods, such as~\cite{efros1999texture,barnes2009patchmatch}, can often perform better (e.g. \reffig{bad_results}).
For semantic inpainting, we compare against nearest neighbor inpainting (which forms the basis of Hays~\etal~\cite{hays2007scene}) and show that our reconstructions are well-aligned semantically, as seen on \reffig{inpaint}.
It also shows that joint loss significantly improves the inpainting over both reconstruction and adversarial loss alone.
Moreover, using our learned features in a nearest-neighbor style inpainting can sometimes improve results over a hand-designed distance metrics.
\reftbl{inpaint_table} reports quantitative results on StreetView Dataset.

\begin{figure}[b]
\vspace{-0.5em}
\centering
\begin{tabular}{c@{\hskip 1pt}c@{\hskip 4pt}c@{\hskip 1pt}c}
\includegraphics[width=0.245\linewidth]{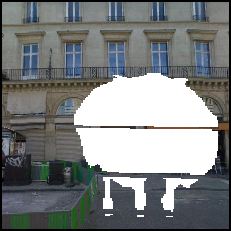} &
\includegraphics[width=0.245\linewidth]{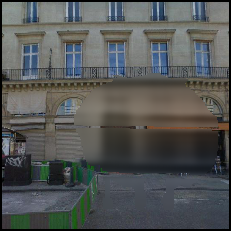} &
\includegraphics[width=0.245\linewidth]{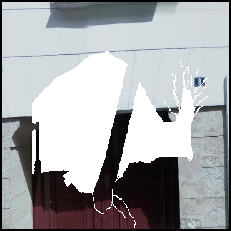} &
\includegraphics[width=0.245\linewidth]{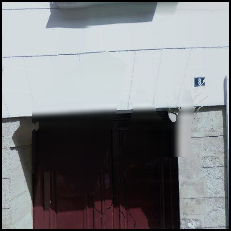}
\end{tabular}
\vspace{-0.5em}
\caption{Arbitrary region inpainting for context encoder trained with reconstruction loss on \textit{held-out} images.}
\lblfig{arbitrary_inpaint}
\vspace{-0.5em}
\end{figure}

\begin{figure*}[t]
\vspace{-0.5em}
\begin{subfigure}[b]{0.5\linewidth}
\centering
\rotatebox{90}{\;\;\;\;Ours}\;\fbox{\includegraphics[width=0.1429\linewidth]{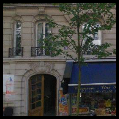}}
\includegraphics[width=0.1429\linewidth]{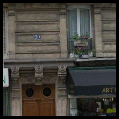}
\includegraphics[width=0.1429\linewidth]{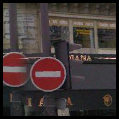}
\includegraphics[width=0.1429\linewidth]{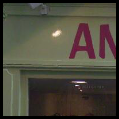}
\includegraphics[width=0.1429\linewidth]{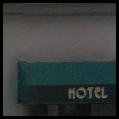}
\includegraphics[width=0.1429\linewidth]{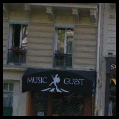}
\end{subfigure}
\begin{subfigure}[b]{0.5\linewidth}
\centering
\rotatebox{90}{\;\;\;\;Ours}\;\fbox{\includegraphics[width=0.1429\linewidth]{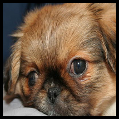}}
\includegraphics[width=0.1429\linewidth]{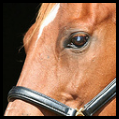}
\includegraphics[width=0.1429\linewidth]{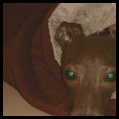}
\includegraphics[width=0.1429\linewidth]{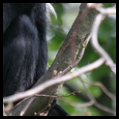}
\includegraphics[width=0.1429\linewidth]{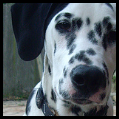}
\includegraphics[width=0.1429\linewidth]{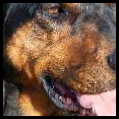}
\end{subfigure}
\begin{subfigure}[b]{0.5\linewidth}
\centering
\rotatebox{90}{\;\;\;\;HOG}\;\fbox{\includegraphics[width=0.1429\linewidth]{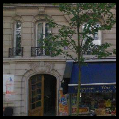}}
\includegraphics[width=0.1429\linewidth]{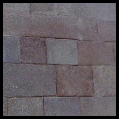}
\includegraphics[width=0.1429\linewidth]{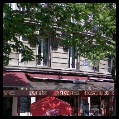}
\includegraphics[width=0.1429\linewidth]{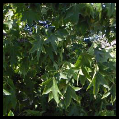}
\includegraphics[width=0.1429\linewidth]{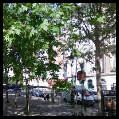}
\includegraphics[width=0.1429\linewidth]{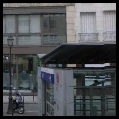}
\end{subfigure}
\begin{subfigure}[b]{0.5\linewidth}
\centering
\rotatebox{90}{\;\;\;\;HOG}\;\fbox{\includegraphics[width=0.1429\linewidth]{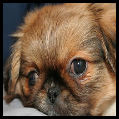}}
\includegraphics[width=0.1429\linewidth]{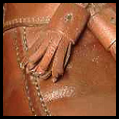}
\includegraphics[width=0.1429\linewidth]{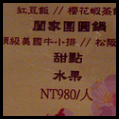}
\includegraphics[width=0.1429\linewidth]{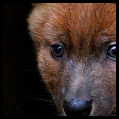}
\includegraphics[width=0.1429\linewidth]{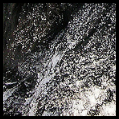}
\includegraphics[width=0.1429\linewidth]{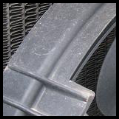}
\end{subfigure}
\begin{subfigure}[b]{0.5\linewidth}
\centering
\rotatebox{90}{\;AlexNet}\;\fbox{\includegraphics[width=0.1429\linewidth]{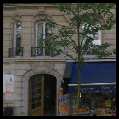}}
\includegraphics[width=0.1429\linewidth]{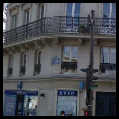}
\includegraphics[width=0.1429\linewidth]{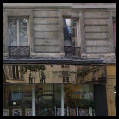}
\includegraphics[width=0.1429\linewidth]{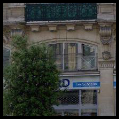}
\includegraphics[width=0.1429\linewidth]{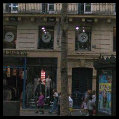}
\includegraphics[width=0.1429\linewidth]{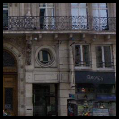}
\end{subfigure}
\begin{subfigure}[b]{0.5\linewidth}
\centering
\rotatebox{90}{\;AlexNet}\;\fbox{\includegraphics[width=0.1429\linewidth]{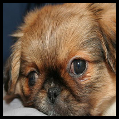}}
\includegraphics[width=0.1429\linewidth]{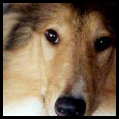}
\includegraphics[width=0.1429\linewidth]{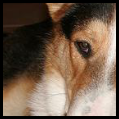}
\includegraphics[width=0.1429\linewidth]{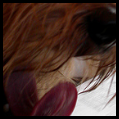}
\includegraphics[width=0.1429\linewidth]{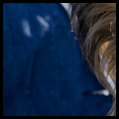}
\includegraphics[width=0.1429\linewidth]{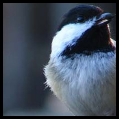}
\end{subfigure}
\vspace{-1em}
\caption{Context Nearest Neighbors. Center patches whose context (not shown here) are close in the embedding space of different methods (namely our context encoder, HOG and AlexNet).
Note that the appearance of these center patches themselves was never seen by these methods.
But our method brings them close just from their context.
}
\lblfig{nnpatches}
\end{figure*}

\begin{table*}[t]
\centering
\begin{tabular}{lccccc}
\toprule
Pretraining Method & Supervision & Pretraining time & Classification & Detection & Segmentation\\
\midrule
ImageNet~\cite{krizhevsky2012imagenet} & $1000$ class labels & $3$ days & 78.2\% & 56.8\% & 48.0\%\\
\midrule
Random Gaussian & initialization & $<1$ minute & 53.3\% & 43.4\% & 19.8\%\\
Autoencoder & - & $14$ hours & 53.8\% & 41.9\% & 25.2\%\\
Agrawal~\etal~\cite{agrawal2015learning} & egomotion & $10$ hours & 52.9\% & 41.8\% & -\\
Wang~\etal~\cite{wang2015unsupervised} & motion & $1$ week & 58.7\% & 47.4\% & -\\
Doersch~\etal~\cite{doersch2015unsupervised} & relative context & $4$ weeks & 55.3\% & 46.6\% & -\\
\midrule
Ours & context & $14$ hours & 56.5\% & 44.5\% & 30.0\%\\
\bottomrule
\end{tabular}
\caption{Quantitative comparison for classification, detection and semantic segmentation. Classification and Fast-RCNN Detection results are on the PASCAL VOC 2007 test set. Semantic segmentation results are on the PASCAL VOC 2012 validation set from the FCN evaluation described in Section~\ref{sec:fcn}, using the additional training data from~\cite{hariharan2011semantic}, and removing overlapping images from the validation set~\cite{long2014fully}.}
\lbltbl{feature_results}
\vspace{-0.5em}
\end{table*}

\subsection{Feature Learning}
\lblsec{feature}
For consistency with prior work, we use the AlexNet~\cite{krizhevsky2012imagenet} architecture for our encoder.
Unfortunately, we did not manage to make the adversarial loss converge with AlexNet, so we used just the reconstruction loss.
The networks were trained with a constant learning rate of $10^{-3}$ for the center-region masks.
However, for random region corruption, we found a learning rate of $10^{-4}$ to perform better.
We apply dropout with a rate of 0.5 just for the channel-wise fully connected layer, since it has more parameters than other layers and might be prone to overfitting.
The training process is fast and converges in about 100K iterations: $14$ hours on a Titan X GPU.
\reffig{arbitrary_inpaint} shows inpainting results for context encoder trained with random region corruption using reconstruction loss.
To evaluate the quality of features, we find nearest neighbors to the masked part of image just by using the features from the context, see~\reffig{nnpatches}.
Note that none of the methods ever see the center part of any image, whether a query or dataset image.
Our features retrieve decent nearest neighbors just from context, even though actual prediction is blurry with L2 loss.
AlexNet features also perform decently as they were trained with 1M labels for semantic tasks, HOG on the other hand fail to get the semantics.

\subsubsection{Classification pre-training}
\label{sec:class}
For this experiment, we fine-tune a standard AlexNet classifier on the PASCAL VOC 2007~\cite{everingham2014pascal} from a number of supervised, self-supervised and unsupervised initializations.
We train the classifier using random cropping, and then evaluate it using $10$ random crops per test image.
We average the classifier output over those random crops.
\reftbl{feature_results} shows the standard mean average precision (mAP) score for all compared methods.

A random initialization performs roughly $25\%$ below an ImageNet-trained model; however, it does not use any labels.
Context encoders are competitive with concurrent self-supervised feature learning methods \cite{doersch2015unsupervised,wang2015unsupervised} and significantly outperform autoencoders and Agrawal~\etal~\cite{agrawal2015learning}.

\subsubsection{Detection pre-training}
\label{sec:detect}
Our second set of quantitative results involves using our features for object detection.
We use \textit{Fast R-CNN}~\cite{fastrcnn} framework (FRCN).
We replace the ImageNet pre-trained network with our context encoders (or any other baseline model).
In particular, we take the pre-trained encoder weights up to the \textit{pool5} layer and re-initialize the fully-connected layers.
We then follow the training and evaluation procedures from FRCN and report the accuracy (in mAP) of the resulting detector.

Our results on the test set of the PASCAL VOC 2007~\cite{everingham2014pascal} detection challenge are reported in~\reftbl{feature_results}.
Context encoder pre-training is competitive with the existing methods achieving significant boost over the baseline.
Recently, Kr\"ahenb\"uhl~\etal~\cite{krahenbuhl2015data} proposed a data-dependent method for rescaling pre-trained model weights.
This significantly improves the features in Doersch~\etal~\cite{doersch2015unsupervised} up to $65.3\%$ for classification and $51.1\%$ for detection.
However, this rescaling doesn't improve results for other methods, including ours.

\subsubsection{Semantic Segmentation pre-training}
\label{sec:fcn}
Our last quantitative evaluation explores the utility of context encoder training for pixel-wise semantic segmentation.
\textit{Fully convolutional networks}~\cite{long2014fully} (FCNs) were proposed as an end-to-end learnable method of predicting a semantic label at each pixel of an image, using a convolutional network pre-trained for ImageNet classification.
We replace the classification pre-trained network used in the FCN method with our context encoders, afterwards following the FCN training and evaluation procedure for direct comparison with their original~\textit{CaffeNet}-based result.

Our results on the PASCAL VOC 2012~\cite{everingham2014pascal} validation set are reported in~\reftbl{feature_results}.
In this setting, we outperform a randomly initialized network as well as a plain autoencoder which is trained simply to reconstruct its full input.

\section{Conclusion}
Our context encoders trained to generate images conditioned on context advance the state of the art in semantic inpainting,
at the same time learn feature representations that are competitive with other models trained with auxiliary supervision.

\paragraph{Acknowledgements}
The authors would like to thank Amanda Buster for the artwork on~\shortreffig{teaser_artist}, as well as Shubham Tulsiani and Saurabh Gupta for helpful discussions. This work was supported in part by DARPA, AFRL, Intel, DoD MURI award N000141110688, NSF awards  IIS-1212798, IIS-1427425, and IIS-1536003, the Berkeley Vision and Learning Center and Berkeley Deep Drive.

{\small
\bibliographystyle{ieee}
\bibliography{main}
}

\section*{Supplementary Material}
In this section, we present the architectural details of our context-encoders, and show additional qualitative results.
Context encoders are not only able to inpaint semantic details in the missing part of an input image, but also learn features transferable to other tasks.
We discuss the implementation details for each of these in following sections.

\subsection*{A. Semantic Inpainting}
Context encoders for inpainting are trained jointly with reconstruction and adversarial loss as discussed in~\refsec{inpaint}.
The inpainting results are slightly worse if we use $227\times 227$ directly.
So, we resize images to $128\times 128$ and then train our joint loss with the resized images.
The encoder and discriminator architecture is similar to that of discriminator in~\cite{dcgan}, and decoder is similar to generator in~\cite{dcgan}; the bottleneck is of $4000$ units.
We used batch normalization in both context encoder and discriminator.
ReLU~\cite{krizhevsky2012imagenet} non-linearity is used in decoder, while leaky ReLU~\cite{dcgan} is used in both encoder and discriminator.

In case of arbitrary region inpainting, adversarial discriminator compares the full real image and the full generated image. We do not condition the adversarial discriminator with mask, see~\eqref{eq:loss_adv}. If the discriminator sees the mask, it figures out the perceptual discontinuity of generated part from the real part and easily classifies the real v/s the generated image, i.e., the process doesn't train.
Moreover, particularly for center region inpainting, this process can be computationally simplified by producing center only and not showing discriminator the context boundary (or in other words, not showing the mask).
The exact architecture for center region dropout is shown in~\reffig{arch_inpaint}.

\subsection*{B. Feature Learning}
We use the AlexNet~\cite{krizhevsky2012imagenet} architecture for encoder so that we can compare the learned features with the prior works, which are trained using Imagenet labels and other un/self-supervised techniques.
The encoder is Alexnet until pool5, followed by channel-wise fully connected layer and decoder is a series of upconvolutional layers until we reach the target size.
The input image size is $227\times 227$.
Unfortunately, we couldn't train adversary with Alexnet Encoder, so it is trained with reconstruction loss.
See~\reffig{arch_feature} for exact architecture details.
For pre-training experiments in~\refsec{feature}, we randomly initialize the fully-connected layers, i.e., \textit{fc}6 and \textit{fc}7, while starting from context encoder weights.

\subsection*{C. Additional Results}
Finally, we show additional inpainting results using our context-encoders in~\reffig{inpaint_supp}.
These results, in comparison to nearest-neighbor inpainting, show that: (a) The features learned by context-encoder are semantically meaningful and retrieve neighboring patches just by looking at the context. This is also verified quantitatively in~\reftbl{feature_results}. (b) Our context encoder doesn't memorize the examples from training set. It rather produces realistic and coherent inpainting results which are much better than nearest neighbor inpainting both qualitatively (\reffig{inpaint_supp}) and quantitatively (\reftbl{inpaint_table}).

\begin{figure*}[t]
\captionsetup[subfigure]{justification=centering}
  \begin{subfigure}[b]{\textwidth}
    \centering
    \includegraphics[width=\textwidth]{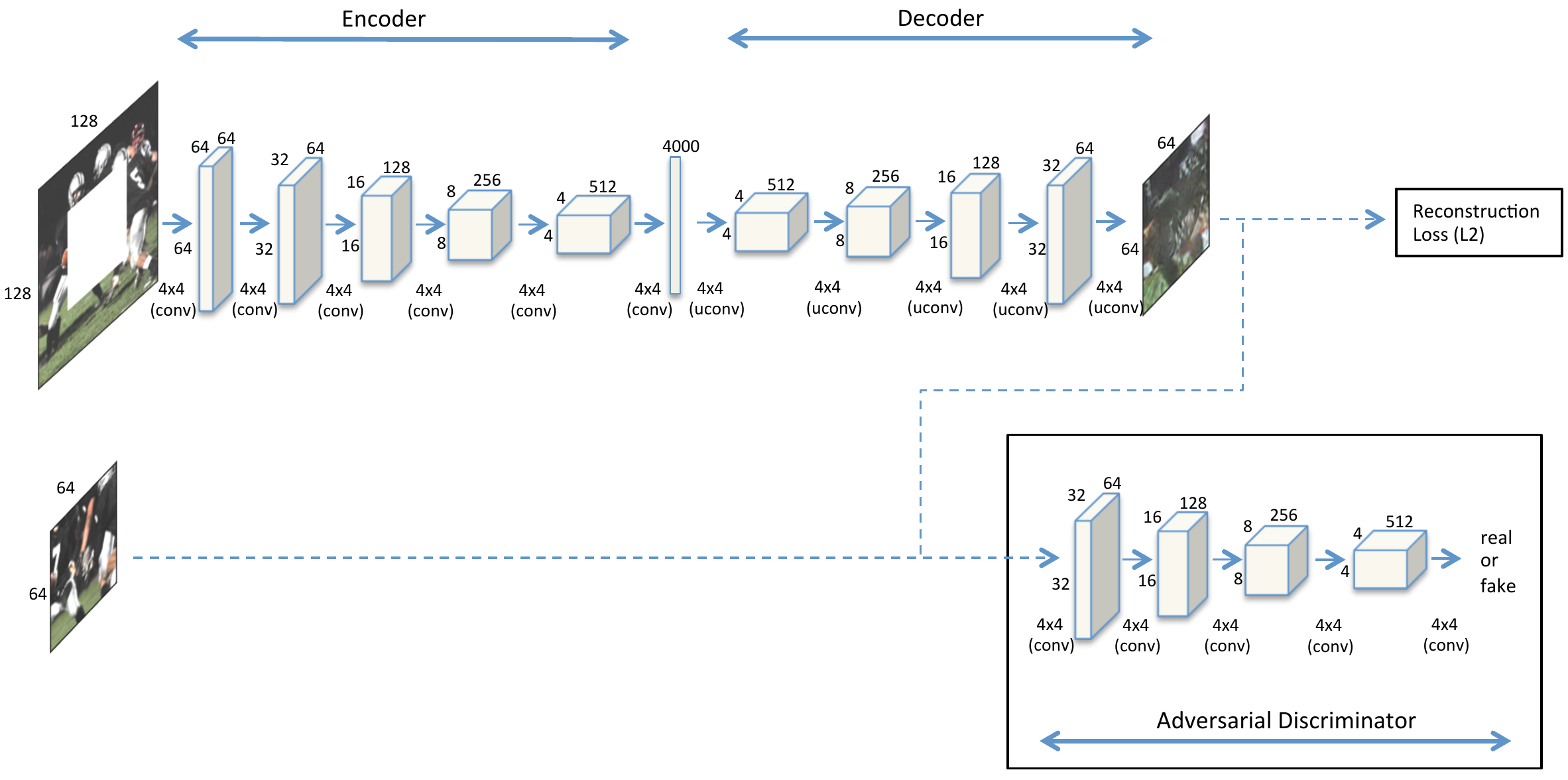}
    \caption{Context encoder trained with joint reconstruction and adversarial loss for semantic inpainting. This illustration is shown for \textit{center region dropout}. Similar architecture holds for arbitrary region dropout as well. See~\refsec{loss}.}
    \lblfig{arch_inpaint}
  \end{subfigure}

  \vspace{4em}
  \begin{subfigure}[b]{\textwidth}
    \centering
    \includegraphics[width=\textwidth]{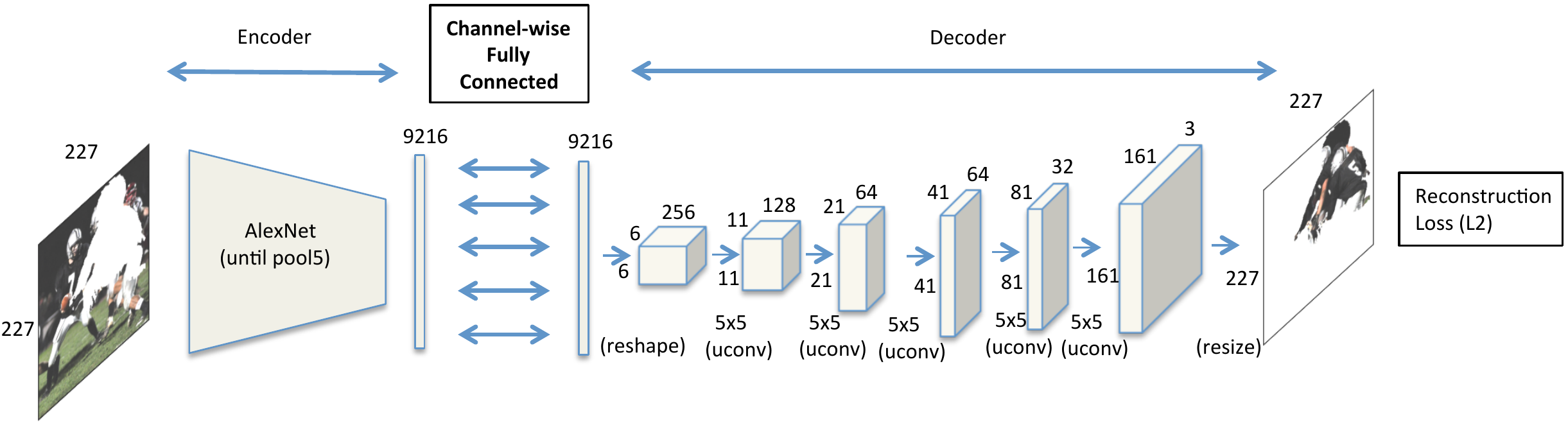}
    \caption{Context encoder trained with reconstruction loss for feature learning by filling in \textit{arbitrary region dropouts} in the input.}
    \lblfig{arch_feature}
  \end{subfigure}
  \vspace{2em}
  \caption{Context encoder training architectures.}
  \lblfig{arch}
\end{figure*}

\begin{figure*}[t]
\centering
\includegraphics[width=\linewidth]{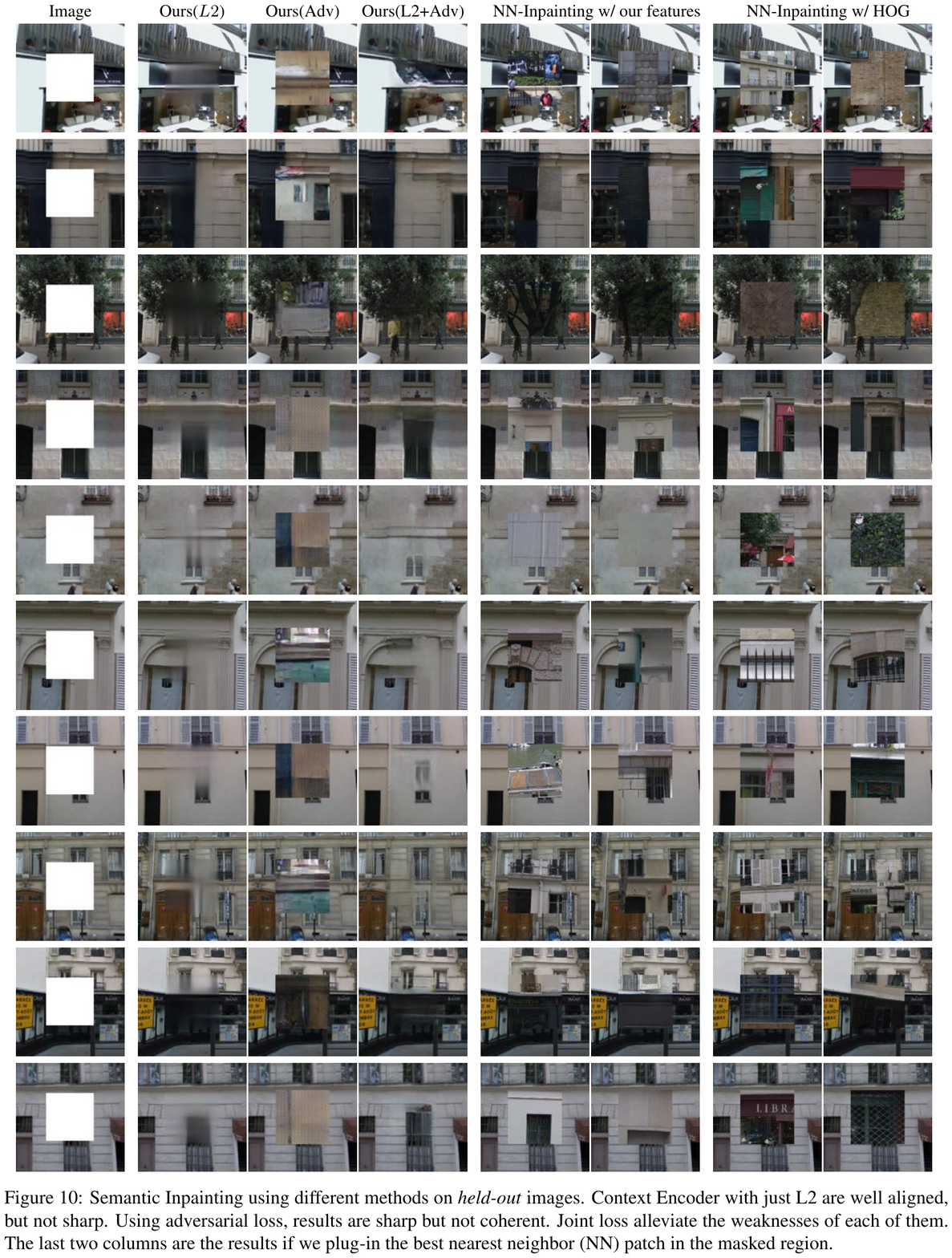}
\caption{Semantic Inpainting using different methods on \textit{held-out} images.
Context Encoder with just L2 are well aligned, but not sharp. Using adversarial loss, results are sharp but not coherent. Joint loss alleviate the weaknesses of each of them.
The last two columns are the results if we plug-in the best nearest neighbor (NN) patch in the masked region.
}
\lblfig{inpaint_supp}
\end{figure*}

\end{document}